\documentclass{article}

\usepackage[nonatbib,preprint]{neurips_2026}

\usepackage[utf8]{inputenc} 
\usepackage[T1]{fontenc}    
\usepackage{hyperref}       
\usepackage{url}            
\usepackage{booktabs}       
\usepackage{amsfonts}       
\usepackage{nicefrac}       
\usepackage{microtype}      
\usepackage[table]{xcolor}         
\usepackage{adjustbox}
\usepackage{subcaption}
\usepackage{amsmath}
\usepackage{wrapfig}
\usepackage{float}
\usepackage{tabularx}

\newcommand{\piold}{\pi_{\text{old}}}

\newcommand{\pitheta}{\pi_{\theta}}

\newcolumntype{H}{>{\setbox0=\hbox\bgroup}c<{\egroup}@{}}
\usepackage{bm}
\title{Guidance Contrastive Token Credit Assignment for Discrete Policy Optimization}

\author{
Shufan Li$^{*1}$, Konstantinos Kallidromitis$^{*2}$, Akash Gokul$^{*3}$ \\ \textbf{Yuta Kyuragi$^2$, Aditya Grover$^1$}
\\ $^1$ UCLA~ $^2$Panasonic AI Research~ $^3$NVIDIA
\\
{ \tt\small *Equal Contribution }
\\
{ \tt\small Correspondence to jacklishufan@cs.ucla.edu}
}

\newcommand{\ours}[0]{GCPO}
\begin{document}

\maketitle

\begin{abstract}
Group-advantage-based reinforcement learning methods, such as GRPO and DAPO, have demonstrated strong performance across diverse domains, including mathematical reasoning and text-to-image generation. However, their reliance on sample-level rewards introduces a key limitation as uniform credit assignment across all tokens fails to capture fine-grained, token-level contributions. To address this issue, we propose Guidance Contrastive Policy Optimization (GCPO), a novel algorithm that enables per-token credit assignment by contrasting model predictions under positive and negative prompts. Rather than uniformly broadcasting sample-level advantages, GCPO assigns token-level advantages proportional to the difference between these contrastive predictions, allowing more precise and informative learning signals. Empirically, we find that GCPO emphasizes semantically relevant regions—such as visual areas aligned with textual prompts in text-to-image generation, and critical keywords within reasoning traces for chain-of-thought tasks. Through extensive experiments, GCPO consistently outperforms GRPO and DAPO baselines on both text-to-image generation and chain-of-thought reasoning benchmarks, demonstrating its effectiveness as a general and scalable optimization strategy for discrete policy learning. Code will be available at \href{https://github.com/jacklishufan/gcpo}{https://github.com/jacklishufan/gcpo}
\end{abstract}

\section{Introduction}

Reinforcement Learning (RL) has proven to be a highly effective post-training approach for generative models in a wide range of scenarios including math reasoning, coding, and text-to-image synthesis \cite{schulman2017proximal, christiano2017deep,rafailov2023direct, deepseek2025r1, yu2025perception,zheng2025group, ouyang2022training,liu2025flow, zheng2025diffusionnft}. Recently, Group Relative Policy Optimization (GRPO) \cite{shao2024deepseekmath} and its variants have been widely adopted to improve models, such as large language models (LLMs), across many domains and problem scales. 

In contrast to prior works like PPO\cite{schulman2017proximal}, which use a learnable value model and generalized advantage estimation (GAE)~\cite{schulman2016gae} to obtain per-token supervision signals through the actor-critic framework, a key innovation of GRPO is to replace GAE with a simpler advantage estimator based on group-normalized per-sample rewards. While this design removes the requirement of an extra value network and improves efficiency, scalability, and stability of training, it broadcasts a uniform sample-level rewards to all tokens in a sequence, which potentially neglects important intra-token differences. 

Intuitively, not all tokens are equally important. In chain-of-thought (CoT)~\cite{wei2022chain} reasoning, some words contribute to the substance of the reasoning traces (e.g., a math calculation) and other words are mere fillers (e.g., punctuation or linking verbs). In text-to-image synthesis, some image regions are directly associated with the entities described in the text prompt while other regions have less association with the prompt. An ideal post-training algorithm should be aware of such differences and weigh tokens differently.

\begin{figure}[t]
    \centering
    \includegraphics[width=0.8\linewidth]{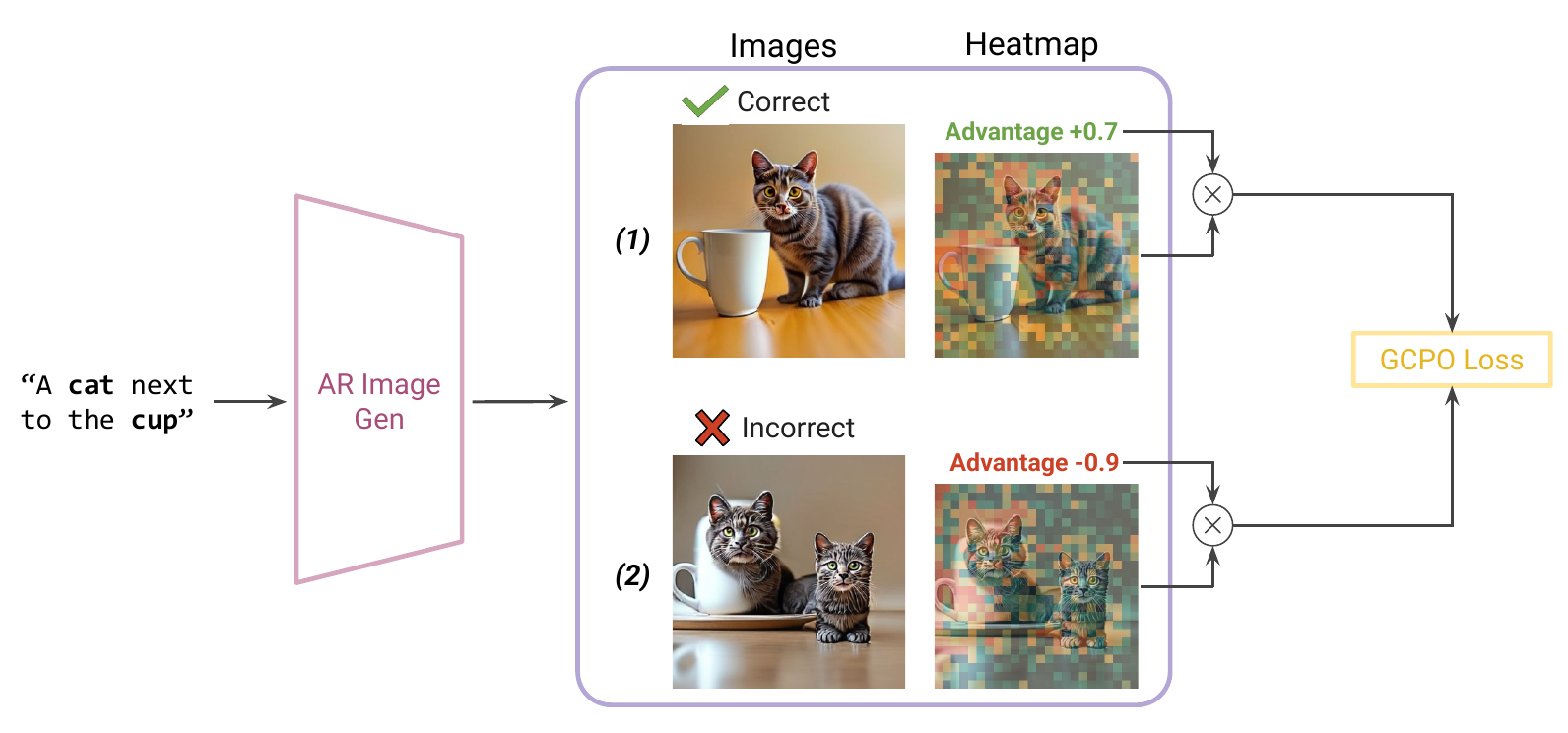}
    \caption{\textbf{GCPO enables fine-grained token credit assignment via contrastive guidance.} GCPO assigns per-token advantages by contrasting the likelihood of each token in a sampled sequence under positive (conditional) and negative (unconditional) prompts. Tokens in semantically critical image regions show high divergence between the two and are amplified, while background regions are down-weighted. The resulting per-token score are used to scale the per-sample advantage in GRPO training.}
    \label{fig:teaser}
\end{figure}

This observation has motivated a growing body of work on token credit assignment, which studies how token-level contributions can be estimated and incorporated into policy optimization objectives. Existing approaches typically construct token-level importance weights or advantages using heuristics derived from quantities such as token confidence, gradient information, entropy, or statistical significance tests \cite{xie2025unlocking, wang2025beyond, sun2025ktae}. More broadly, these methods explore different mechanisms for decomposing sequence-level rewards into finer-grained optimization signals for autoregressive generation, while recent works have also extended token credit assignment to multimodal settings by emphasizing visually grounded tokens \cite{huang2025spotlight}.


In this work, we propose guidance contrastive policy optimization (GCPO), a variant of GRPO that assigns per-token importance weighting by contrasting positive and negative predictions from the same policy. GCPO is largely motivated by classifier-free guidance \cite{ho2022classifier}, a widely adapted inference technique for text-to-image models. CFG augments the model's original prediction by the differences between its prediction with a positive prompt (i.e. user input) and a negative prompt (e.g. empty string). CFG has been shown to improve image quality and text–image alignment in diffusion-based visual generation models, and has become a standard component in modern text-to-image systems \cite{ho2022classifier, nichol2021glide, saharia2022imagen, rombach2022highresolution}. We observe that in discrete image generation, where an image is represented as a sequence of discrete tokens as opposed to continuous latents, not all tokens are affected by CFG equally. In particular, regions that are directly associated with the prompt will have a larger difference between the predictions given a positive and negative prompt, while regions that do not contain entities specified by the prompt are less sensitive. Based on this observation, we build a token credit assignment scheme that assigns high credit to tokens more sensitive to CFG as opposed to broadcasting the sample-level reward uniformly. Specifically, this credit assignment is computed using the KL divergence between per token predictions with positive prompts and with negative prompts. This allows the learning process to focus on key regions in the sampled image, improving training stability and achieving better downstream performance. 

While we draw inspiration from the text-to-image domain, we do not constrain GCPO to visual synthesis. Although less commonly studied, classifier-free guidance has been explored in language modeling as an inference-time control mechanism, where it has been shown to improve reasoning \cite{sanchez2023stay}. More broadly, related contrastive inference methods in LLMs demonstrate that differences between positive and negative conditioning signals can provide meaningful steering signals for generation \cite{o2023contrastive, oneill2023steer}, suggesting that such signals may also be useful for policy learning. Motivated by this observation, we extend GCPO to multimodal chain-of-thought reasoning in vision-language models (VLMs) by incorporating analogous importance weighting over text tokens.

Compared with text-to-image use cases, extending GCPO to VLMs introduces two challenges. First, unlike text-to-image models that naturally support CFG by incorporating unconditional generation tasks via prompt dropping during training, instruction-tuned LLMs always expect a prompt. We explore several options of negative prompts and discovered that simply augmenting the original prompt with the extra instruction ``please generate a wrong answer" at the end works well. Second, while existing works demonstrated CFG can work for VLM inference in certain cases, the common inference protocol of VLMs do not employ CFG because it does not reliably improve model's performance in a predicted manner. Luckily, through empirical observation we find that CFG-inspired token weighting in GCPO still improves the model's performance even though the rollout and sampling process does not explicit employ CFG as an inference technique. 

Another challenge common in both image and text generation is finding a good normalization technique for importance weights. Since the unnormalized KL divergence lies in the range of $(0,\infty)$, naively applying a softmax normalization or min-max normalization will concentrate the probability mass to only a few tokens. Most notably, we observe that the first token in the response commonly exhibits a large KL divergence. A common approach to stabilize these weights is clipping and temperature based scaling. However, we observe that the KL scale is highly dynamic across different tasks, prompts, and sequence lengths, making hyperparameter tuning challenging. To remove the need of task-specific hyperparameter tuning while ensuring a smooth token credit assignment, we employ a rank-based normalization technique that normalizes the per-token KL divergence based on the ranking in the sequence. For example, tokens at the 90th percentile of different sequences will always be normalized to the same value regardless of their absolute scale in their respective sequences. This technique is equivalent to applying a histogram equalization on a heatmap of per-token KL divergence, ensuring roughly identical distribution of normalized weights in each sample. 

Through extensive experiments, we show that GCPO outperforms GRPO and DAPO baselines on both text-to-image generation and multimodal reasoning benchmarks, including GenEval \cite{ghosh2023geneval}, MathVerse \cite{zhang2024mathverse}, MathVision \cite{wang2024measuring}, LogicVista \cite{xiao2024logicvista}, and MMMU-Pro \cite{yue2025mmmu}. These results demonstrate that GCPO is a simple and effective strategy for multimodal understanding and generation tasks.

\section{Background and Related Works}

\subsection{Group-advantaged based reinforcement learning}

Reinforcement learning has proven to be an effective post-training approach for generative models. Early works primarily adopt proximal policy optimization (PPO) \cite{schulman2017proximal, ouyang2022training, stiennon2020learning} within an actor-critic framework, leveraging learned reward models for alignment. More recent approaches such as Direct Preference Optimization (DPO)\cite{rafailov2023direct, Wallace2023DiffusionMA} provide a simpler alternative to reinforcement learning from human feedback by directly optimizing preferences without an explicit reward model or PPO-based training. At the same time, RL-based methods continue to scale effectively in large language and multimodal settings. GRPO \cite{shao2024deepseekmath} removes the need for a learnable value model via group-based advantage estimation, demonstrating strong scalability and effectiveness across domains including mathematical reasoning \cite{shao2024deepseekmath, deepseek2025r1}, multimodal reasoning \cite{shen2025vlm, wang2025vl}, and text-to-image generation \cite{liu2025flow, pan2025janus, ye2025data, zheng2025diffusionnft}. 

Concretely, given a policy model $\pi_\theta$, GRPO maximizes the following training objective.
\begin{align}
\label{equ:grpo}
\mathcal{J}_\text{GRPO}(\theta) = \mathbb{E}
\left[ \frac{1}{N} \sum_{i=1}^{N} \frac{1}{|y_i|} \sum_{t=1}^{|y_i|} 
\min \left( w_{i,t}(\theta) \widehat{A}_{i,t}, \, \mathrm{clip} \left( w_{i,t}(\theta), 1 - {\varepsilon}, 1 + {\varepsilon}\right) \widehat{A}_{i,t} \right)
\right] \\
    w_{i,t}(\theta)=\frac{ \pi_{\theta} (y_{i,t} | x, y_{i,<t}) }{ \pi_{\theta_\text{old}} (y_{i,t} | x,y_{i,<t})},\quad
    \widehat{A}_{i,t} = \widehat{A}_{i} = \frac{r(x, y_i) - \mathrm{mean} \left( \{ r(x, y_i) \}_{i=1}^G \right) }{ \mathrm{std} \left( \{ r(x, y_i) \}_{i=1}^G \right) },
\end{align}
where $y_1, \dots, y_N$ is a group of $N$ responses sampled from the model corresponding to prompt $x$, $r(x,y_i)$ is the per-sample reward function, $\widehat{A}_i$ is the per-sample advantage (i.e. normalized reward), and $w_{i,t}$ is an important weight for trust-region updates. In practice, an additional KL penalty term is often added to $\mathcal{J}_\text{GRPO}$ for training stability. 

Several followup works explored various ways to improve the vanilla GRPO formulation. GSPO \cite{zheng2025group} improves training stability by using a sequence level ratio instead of per-token importance weights. DAPO \cite{yu2025dapo} removes the KL penalty and incorporates asymmetric clipping and online sample filtering. Dr. GRPO \cite{liu2025understanding} removes the standard-deviation based scaling in advantage scaling. These modifications showed varying levels of improvement compared to the GRPO baseline.

\subsection{Token Credit Assignment}

While an importance weight $w_{i,t}$ is incorporated for each token, it only weights the tokens based on their deviations from the trust region by referring to an earlier checkpoint $\piold$, as opposed to the semantic importance of each token. In fully on-policy learning setting, $\piold=\pitheta$ and $w_{i,t}$ is always 1, treating all tokens equally. A specific line of works known as token credit assignment argues that this form of importance weight is insufficient. Conceptually, not all tokens are equally important in terms of their contribution to the final reward. For instances, in math reasoning tasks some tokens representative the substance of reasoning process (e.g. math derivations), while other tokens are mere fillers (e.g. punctuations, linking words). However, the per-token advantage $\widehat{A}_{i,t} = \widehat{A}_{i}$ stays the same for all tokens in a sequence, neglecting such differences. Token credit assignment address this issue by designing algorithms that assigns non-uniform advantages at token level, allowing the training process to focus on important tokens. UCAS \cite{xie2025unlocking} scales advantages based on per-token confidence. OAR scales advantages based on model gradients. Wang et. el. \cite{wang2025beyond} emphasizes advantage signals on high-entropy minority tokens in math reasoning tasks. KATE \cite{sun2025ktae} assigns per-token advantage using p-values of Fisher's exact test. Most of these works are confined to language only space. VPPO \cite{huang2025spotlight} first explored token credit assignment in multimodal understanding setting by emphasizing visually dependent tokens. Token credit assignment in text-to-image generation remains largely unexplored.
\begin{figure}[t]
    \centering
    \includegraphics[width=1\linewidth]{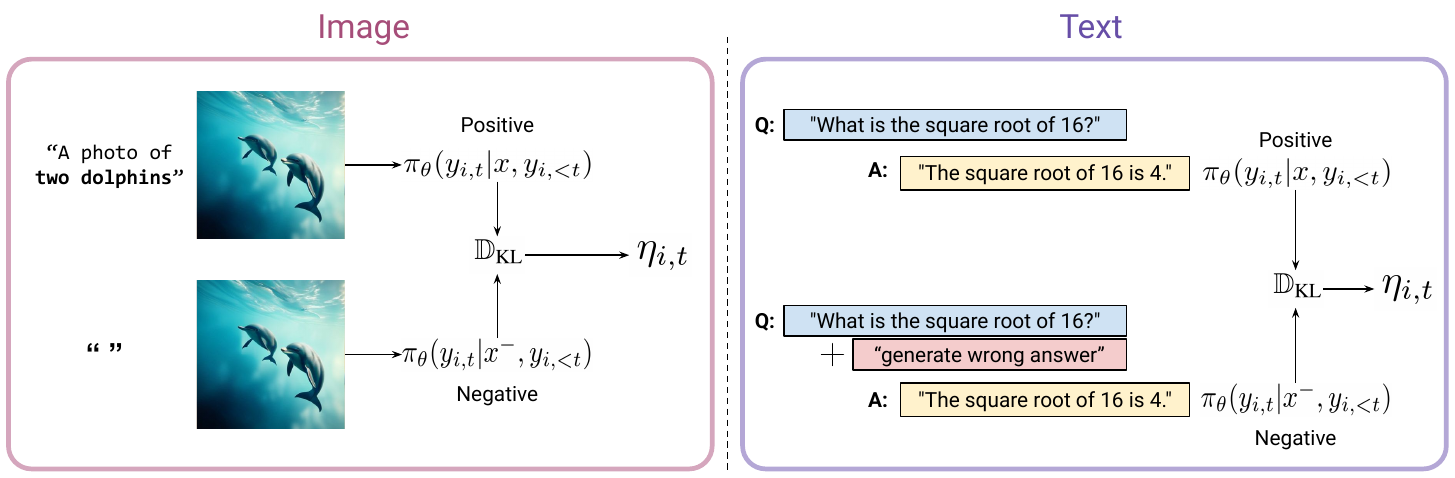}
    \caption{\textbf{GCPO computes per-token importance weights via contrastive guidance.} For text-to-image generation (left), a sampled image sequence is scored by $\pi_{\theta}$ under both positive (conditional, text prompt) and negative (unconditional, empty prompt) inputs, like in CFG. For multimodal reasoning (right), a sampled response is scored under positive (original question) and negative ("generate the wrong answer") prompts. In both cases, the same sequence is scored twice under different prompt conditions, and the per-token KL divergence between the two distributions serves as the basis for per-token advantage weighting in GCPO.}
    \label{fig:explainer}
\end{figure}

\subsection{Classifier-Free Guidance}

Classifier-Free Guidance \cite{ho2022classifier} (CFG) is an inference technique first developed for continuous diffusion model and has inspired similar inference-time guidance mechanisms in autoregressive models. Formally, given an input prompt $x$, we first construct a negative prompt \cite{ban2024understanding}$x^-$. Common choices in text-to-image generation are empty strings and negative keywords such as ``bad quality, deformed" \cite{ban2024understanding}. The model's predicted pre-softmax logits $l_{\theta} (y_{i,t} | x, y_{i,<t})$ is modified by the following formula:
\begin{align}
     l^{\text{CFG}}_\theta (y_{i,t} | x, y_{i,<t})= l_{\theta} (y_{i,t} | x, y_{i,<t})+\lambda(\pi_{\theta} (y_{i,t} | x, y_{i,<t})-l_{\theta} (y_{i,t} | x^-, y_{i,<t}))
\end{align}
where $\lambda$ is the guidance scale. In the language domain, some works explore incorporating CFG-style guidance at inference time for LLMs \cite{sanchez2023stay}. However, CFG is not widely adopted in the inference pipeline of large language models. In this work, we extend CFG into the RL posttraining of autoregressive models, including large language models and text-to-image generations.

\section{Method}

In this work, we explore a novel token credit assignment algorithm based on classifier-free guidance. During text-to-image inference, we observe that not all tokens of an image are affected equally by CFG. Visual tokens that are most affected by CFG tend to be image regions strongly associated with the text prompt (Fig \ref{fig:normalization}). Based on this observation, we propose Guidance Contrastive Policy Optimization (GCPO), that emphasizes learning signals in regions affected most by CFG.

\subsection{Contrastive Guidance}

\begin{figure}
    \centering
    \includegraphics[width=0.5\linewidth]{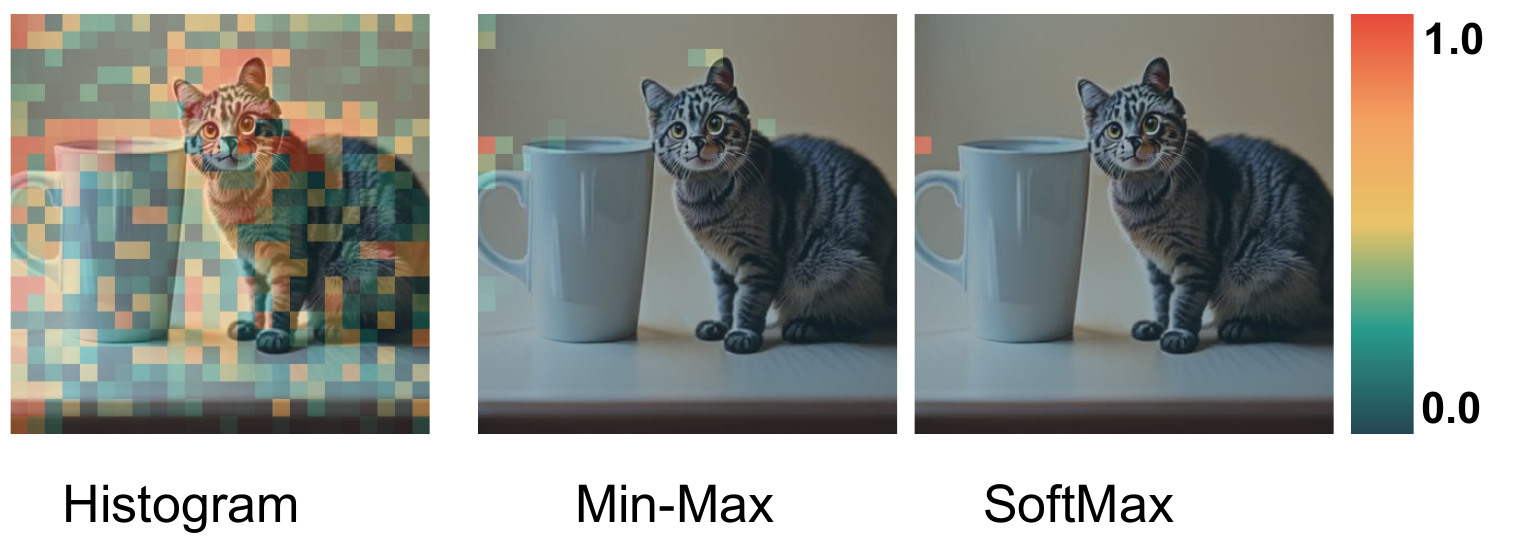}
    \caption{\textbf{Comparison of normalization strategies.} Min-max and softmax are sensitive to outliers. Histogram equalization produces a smooth distribution of weights regardless of the absolute scale, ensuring consistent optimization behavior across different samples and prompts.}
    \label{fig:normalization}
\end{figure}

Formally, given a text prompt $x$ and a generated sequence of image tokens $y_i$. For each token $y_{i,t}$, we can compute the positive probability $\pi_{\theta} (y_{i,t} | x, y_{i,<t})$ and negative probability $\pi_{\theta} (y_{i,t} | x^-, y_{i,<t})$, where $x^-$ is the negative prompt. We then compute the KL divergence of these two distributions to obtain contrastive guidance $\eta$:
\begin{equation}
    \eta_{i,t}=\mathbb{D}_\text{KL}( \pi_{\theta} (y_{i,t} | x, y_{i,<t})||\pi_{\theta} (y_{i,t} | x^-, y_{i,<t}) )
\end{equation}
Intuitively, $\eta\in(0,\infty)$ quantitatively measures the difference between the positive distribution and negative distribution, and is correlated with how much the per-token prediction should be affected by CFG during inference. When $\eta$ is small, the two distributions are close, making the guidance signal less pronounced. In contrast, when $\eta$ is large, the positive distribution and the negative distribution differ drastically, leading to a more pronounced influence of CFG. We empirically observe this in Figure \ref{fig:normalization}, where $\eta$ is high in regions containing the entity described in the prompt and $\eta$ is small in low-frequency background regions. 

\subsection{Normalization}

To convert unbounded $\eta$ to useful weights, we need to normalize $\eta$ into the range of [0,1]. The two most common choices for such normalization are softmax normalization and min-max normalization. However, both choices are suboptimal for our use case because we observe that the raw KL divergence has a huge variance in scales across tokens, prompts, and generated images, making min-max normalization impractical and tuning the temperature of softmax normalization challenging. To address this issue, we propose using histogram equalization, a common image processing technique to normalize the heatmap and ensure a smooth distribution of normalized weights.

Specifically, given unnormalized KL divergence $\eta$, we make use of the cumulative distribution function $F$ to normalize the KL divergence as follows
\begin{align}
    \eta_{\text{normalized}}=F(\eta)=\mathbb{P}(x;x<\eta)
\end{align}
where $\mathbb{P}(x;x<\eta)$ is the percentage of tokens from the same image that is smaller than $\eta$. In this setup, the highest KL value always normalize to 1, the lowest KL value always normalize to 0, the median values always normalize to 0.5, and so forth. Compared with alternatives, this approach ensures equal distribution patterns of normalized weights with each image sequence, leading to smoother optimization process.

\subsection{Guidance Contrastive Policy Optimization}

With normalized weight $\eta_{\text{normalized}}$, we can obtain per token advantages by scaling the per sample advantage $\widehat{A}_i$, Specifically, instead of broadcasting the per-sample advantage $\widehat{A}_i$ to all tokens uniformly as in standard GRPO described in Equation \ref{equ:grpo}, we obtain per-token advantage $\widehat{A}_{i,t}$ by
\begin{align}
    \widehat{A}^{GCPO}_{i,t} &=\eta_{\text{normalized},i,t}\widehat{A}_i \\
     &=\eta_{\text{normalized},i,t} \frac{r(x, y_i) - \mathrm{mean} \left( \{ r(x, y_i) \}_{i=1}^G \right) }{ \mathrm{std} \left( \{ r(x, y_i) \}_{i=1}^G \right) }
\end{align}
The final GCPO objective is as follows:
\begin{align}
\label{equ:gcpo}
\mathcal{J}_\text{GCPO}(\theta) = \mathbb{E}
\left[ \frac{1}{N} \sum_{i=1}^{N} \frac{1}{|y_i|} \sum_{t=1}^{|y_i|} 
\min \left( w_{i,t}(\theta) \widehat{A}^{GCPO}_{i,t}, \, \mathrm{clip} \left( w_{i,t}(\theta), 1 - {\varepsilon}, 1 + {\varepsilon}\right) \widehat{A}^{GCPO}_{i,t} \right)
\right]
\end{align}

Unlike vanilla GRPO, GCPO assigns higher weight to key tokens whose logits are more affected by CFG, tailoring the training process to focus on important regions.

\subsection{Extending to language generation. }

For text-to-image models, CFG is a commonly used inference technique, we can naturally select the negative prompt that is used during the sampling process (e.g. empty string). While CFG is less common in language domain, prior works demonstrated its feasibility \cite{sanchez2023stay}. We explored extending GCPO to language generation in multimodal reasoning tasks. 

Given input $x$ consisting of input images and text instructions, we sample $N$ response $y_1...y_N$ from a VLM. Unlike the image generation setup, we do not employ CFG during rollout to align with prior works on multimodal reasoning. However, we can still compute divergences between $\pi_{\theta} (y_{i,t} | x, y_{i,<t}$ and $\pi_{\theta} (y_{i,t} | x^-, y_{i,<t})$ to find tokens that are most affected by a hypothetical CFG.

Since instruction-tuned VLMs do not naturally accept empty strings as an instruction, we need to manually construct a negative prompts $x^-$. In this work, we explored several options including a generic prompt "answering the question" and augmenting the original question with a suffix instruction "give the wrong answer to this question", we find that the latter works best in practice.

Conceptually, this choice of $x^-$ has a specific Bayesian interpretation. Given a prompt $x$ and response $y$, the probability $\pi_{\theta} (y | x)$ is implicitly the probability of a correct answer $\pi_{\theta} (y| x, \text{correct})$ in a reasonably trained language model. In our construct, $x^-$ appends the suffix and an instruction which tells the model to predict the wrong answer, this amounts to $\pi_{\theta} (y| x, \text{incorrect})$. Through Bayesian relations, we can derive the following relation
\begin{align}
    \frac{\pi_{\theta} (y | x)}{\pi_{\theta} (y | x^-)} = \frac{\pi_{\theta} (y | x,\text{correct})}{\pi_{\theta} (y | x,\text{incorrect})}=\frac{\pi_{\theta}(\text{correct}|x,y)}{\pi_{\theta}(\text{incorrect}|x,y)} \frac{\pi_{\theta}(x,\text{incorrect})}{\pi_{\theta}(x,\text{correct})}
    \label{eq:odds}
\end{align}
where $\frac{\pi_{\theta}(x,\text{incorrect})}{\pi_{\theta}(x,\text{correct})}$ dependents only on the prompt $x$ and is constant for all corresponding responses. The term $\frac{\pi_{\theta}(\text{correct}|x,y)}{\pi_{\theta}(\text{incorrect}|x,y)}$ is the odds of an implicit classifier that reflects the model's belief in the correctness of $y$. On the token level, this can be interpreted as the odds of an implicit classifier that determines if a token belongs to a correct answer and incorrect answer. When the odds are very small or very large, it indicates that a token is a high probable correct or incorrect token. The KL divergence term in GCPO naturally assign large weights to these tokens.


\textbf{Connection with VPPO}. GCPO is closely related to recent work on VPPO \cite{huang2025spotlight} on multimodal reasoning, which tailors the reinforcement learning signals for VLMs to visually dependent tokens. We note that VPPO is confined to visually grounded tasks while GCPO is motivated by classifier free guidance and has broader applications. For the specific task of multimodal reasoning, VPPO can be considered as a variant of GCPO with following design choices: First, it constructs $x^-$ by randomly masking part of the input images as opposed to adding the suffix ``generate the wrong answer". Second, it uses a hard filter and set the importance weight to 1 for top 40\% of visually dependent tokens while set the weight of other tokens to 0. We argue GCPO is superior because while visually dependent tokens are important, multimodal reasoning tasks also requires generally reasoning capabilities. For example, when solving geometry problems, being able to understand the shapes and labels in the input image is important, but correctly applying math derivation and numerical computations are equally important. VPPO's choice of negative prompts and its use of hard filter limit the learning signal in these important aspects. We argue GCPO is more preferable than VPPO even for multimodal reasoning tasks, and empirically validate this through experiments. We defer additional discussions to the experiment section.

\section{Experiments}

\subsection{Text-to-Image Generation}
To validate the effectiveness of GCPO, we first conducted text-to-image generation experiments on an autoregressive image generation model Janus-Pro-7B. We use the GRPO as the main baseline and also compare with a previous RL methods on the same model Janus-Pro-R1 \cite{pan2025janus}, as well as state-of-the-art text-to-image models like FLUX.1-dev \cite{flux2024}, Stable Diffusion 3 \cite{esser2024scaling-sd3}, and Qwen-Image \cite{wu2025qwen}. We use the training data of FlowGRPO \cite{liu2025flow} and its implementation of GenEval reward model, which provide verifiable reward for text-to-image generation tasks by checking if the generated images matches the prompt specification via object detectors and classification models. We provide additional implementation details such as learning rate and optimizer schedules in Appendix A. 

\begin{figure}[t]
    \centering
    \includegraphics[width=1\linewidth]{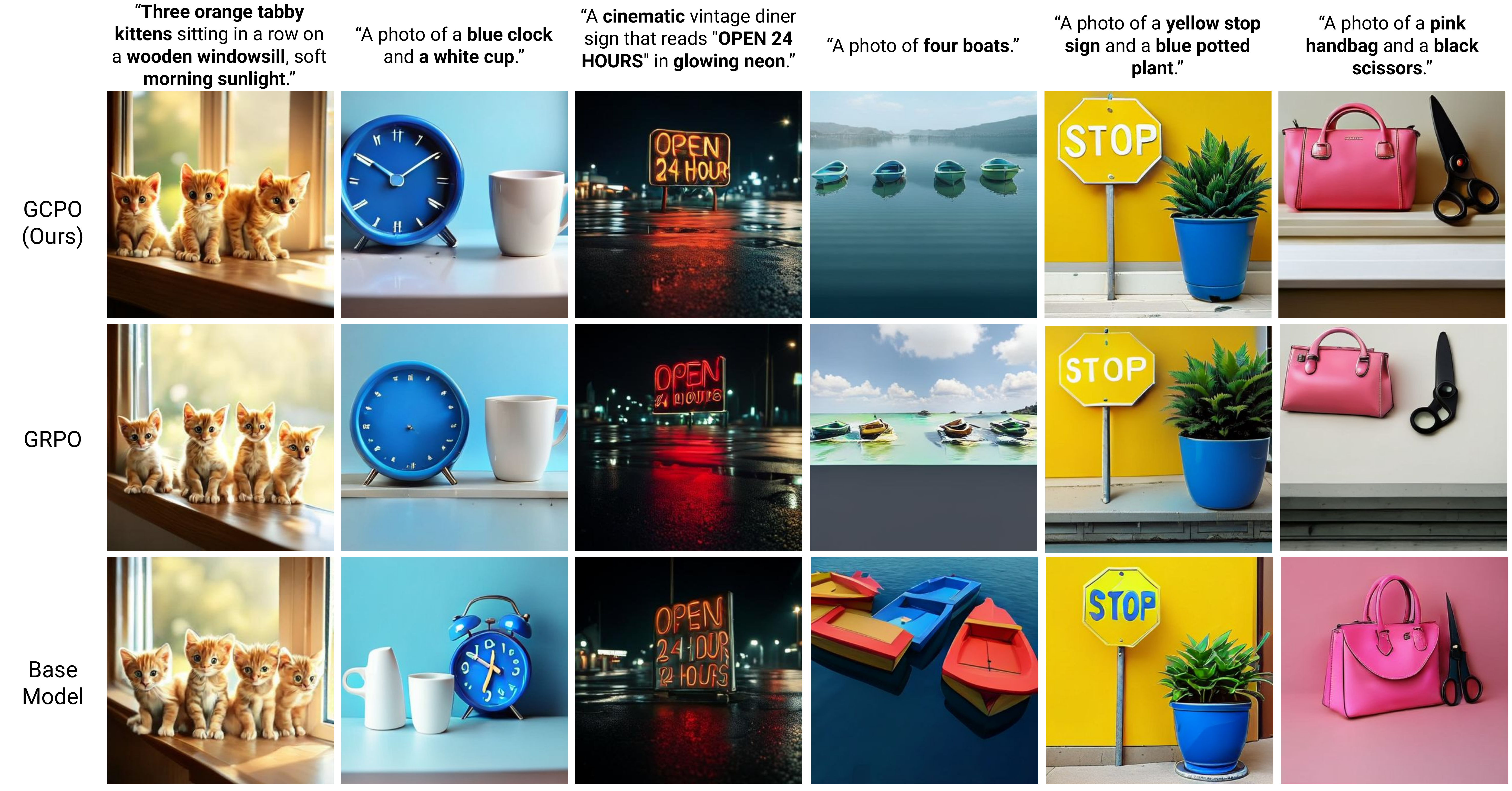}
    \caption{\textbf{Comparison of GCPO vs GRPO vs Base Model (Janus-Pro-7B).} We evaluate the models on GenEval to observe quality and text conditioning. GCPO is able to produce higher quality outputs (Clock, Boats, Scissors) that are more consistent with the text instructions ("apple above tv").}
    \label{fig:side_by_side}
    
\end{figure}

We report GenEval benchmark scores in Table \ref{tab:gen_eval_results}. GCPO achieves high performance with an overall score of 0.89, which improves from the Janus-Pro-7B base model by (+0.09). It outperforms a previous RL method Janus-Pro-R1 derived from the same base model, as well as the GRPO baseline, highlighting the effectiveness of GCPO. Compared with other models, GCPO achieves comparable performance to state-of-the art image generators such as Qwen-Image-2507 and BAGEL, which are significantly larger in terms of parameters. Among subcategories, the performance gain is most pronounced in the counting ($0.56\rightarrow0.84$) and color attribution ($0.66\rightarrow0.83$), which naturally benefit from GCPO's importance weight which focus key regions. In additional to quantitative results, we also provide qualitative comparisons in Figure \ref{fig:side_by_side}. Images generated with GCPO tuned model better represents user prompts.

\begin{table*}[t!]
  \centering
  \caption{\textbf{GenEval benchmark results for text-to-image generation across state-of-the-art models.} Janus-Pro-R1 + GCPO achieve the highest overall score (0.89), outperforming same-size baselines and matching significantly larger models.}
  \label{tab:gen_eval_results}
  \begin{adjustbox}{max width=\textwidth}
  \begin{tabular}{ccccccccc}
    \toprule
    \textbf{Model} & \textbf{Params} & \textbf{Single Obj.$\uparrow$} & \textbf{Two Obj.$\uparrow$} & \textbf{Counting$\uparrow$} & \textbf{Colors$\uparrow$} & \textbf{Position$\uparrow$} & \textbf{Color Attri.$\uparrow$} & \textbf{Overall$\uparrow$} \\

    \midrule
    Emu3~\cite{wang2024emu3} & 8B & - & - & - & - & - & - & 0.66  \\
    Janus-Pro~\cite{chen2025janus} & 7B & 0.99 & 0.89 & 0.59 & 0.90 & 0.79 & 0.66 & 0.80   \\
    MMaDA~\cite{yang2025mmada} & 8B & 0.99 & 0.76 & 0.61 & 0.84 & 0.20 & 0.37 & 0.63  \\
    Show-o~\cite{xie2025show} & 1.3B & 0.98 & 0.80 & 0.66 & 0.84 & 0.31 & 0.50 & 0.68 \\
    BAGEL~\cite{deng2025emerging} & 14B & 0.98 & 0.95 & 0.84 & 0.95 & 0.78 & 0.77 & 0.88 \\
    LaViDa-O~\cite{li2025lavidao} & 10B & 0.99 & 0.85 & 0.71 & 0.86 & 0.65 & 0.58 & 0.77  \\

    Show-o2~\cite{xie2025show} & 7B & 1.00 & 0.87 & 0.58 & 0.92 & 0.52 & 0.62 & 0.76  \\
    PixArt-$\alpha$~\cite{chen2023pixart} & 0.6B & 0.98 & 0.50 & 0.44 & 0.80 & 0.08 & 0.07 & 0.48 \\
    DALL-E 3~\cite{openai_dalle3} & - & 0.96 & 0.87 & 0.47 & 0.83 & 0.43 & 0.45 & 0.67 \\
    SD3-Medium~\cite{esser2024scaling-sd3} & 2B & 0.99 & 0.94 & 0.72 & 0.89 & 0.33 & 0.60 & 0.74  \\
    FLUX.1-dev~\cite{flux2024} & 12B & 0.98 & 0.81 & 0.74 & 0.79 & 0.22 & 0.45 & 0.66  \\
    Qwen-Image-2507 \cite{wu2025qwen} & 20B & 0.99 & 0.92&  0.89&  0.88&  0.76 & 0.77& 0.87  \\  
    \midrule
        Janus-Pro~\cite{chen2025janus} & 7B & 0.99 & 0.89 & 0.59 & 0.90 & 0.79 & 0.66 & 0.80   \\
        + Janus-Pro-R1 \cite{pan2025janus} &   7B  &  0.99& 0.94& 0.66 &0.92 &0.87 & 0.78 & 0.86 \\
        + GRPO $\dagger$   &   7B    & 0.99 & 0.93 & 0.81 & 0.83 & 0.83 & 0.73 & 0.85\\ 
        \rowcolor{gray!20}
          + GCPO    &   7B    & 1.00 & 0.95 & 0.84 & 0.89 & 0.83 & 0.83 & 0.89 \\ 
    \bottomrule
  \end{tabular}
  \end{adjustbox}
\end{table*}

\subsection{Multimodal-reasoning}

\begin{table}[t]
\centering
\caption{\textbf{Performance comparison across multimodal reasoning benchmarks.} All models are trained on ViRL-39k dataset. $\dagger$ results are cited from VPPO \cite{huang2025spotlight}.}
\label{tab:mm_results}
\begin{tabular}{lccccc}
\toprule
Method & MathVerse & MathVision & MM12k & LogicVista & MMMU-Pro \\
\midrule
\quad Qwen2.5-VL-Instruct$\dagger$ & 39.0 & 18.4 & 42.5 & 42.4 & 25.1 \\
\quad +GRPO$\dagger$    & 66.5 & 30.7 & 72.3 & 45.6 & 35.2 \\
\quad +DAPO$\dagger$    & 68.3 & 30.5 & 82.1 & 46.8 & 35.9 \\
\quad +VPPO$\dagger$    & 71.6 & 33.3 & 82.8 & 47.9 & 37.9 \\
\rowcolor{gray!20}
\quad +GCPO    & 71.7 & 31.6 & 83.1 & 49.7 & 39.3 \\
\midrule
\quad Qwen3-VL-Instruct$\dagger$ & 67.7 & 37.8 & 68.8 & 50.1 & 33.7 \\
\quad +DAPO    & 76.5 & 44.4 & 75.1 & 60.0 & 50.5 \\
\quad +VPPO$\dagger$    & 83.8 & 55.9 & 80.6 & 62.3 & 52.4 \\
\rowcolor{gray!20}
\quad +GCPO    & 84.1 & 56.9 & 81.3 & 63.0 & 55.3 \\
\bottomrule
\end{tabular}
\end{table}

We extend GCPO to language generation tasks in the context of multimodal reasoning of VLMs. We adopt the setting of a prior work VPPO \cite{huang2025spotlight} and also compare with GRPO and DAPO baselines. We use Qwen2.5-VL-Instruct-7B \cite{bai2025qwen25-vl} and Qwen3-VL-Instruct-8B \cite{Qwen3-VL} as the base model and perform reinforcement learning with correctness reward of ViRL39K \cite{wang2025vl} dataset. Following VPPO, we build our method on top of DAPO instead of GRPO because it leads to stronger performance and better stability. In particular, we find that for Qwen3-VL-Instruct-8B model, vanilla GRPO easily divergences after 20 steps, while the online sample filtering technique of DAPO helps stabilized the training by remove groups with only correct answers and groups with only incorrect answers. We provide additional training details in the Appendix A.

We report results on visual math reasoning benchmarks including MathVerse \cite{zhang2024mathverse}, MathVision \cite{wang2024measuring}, and the test split of MM12K\cite{meng2025mm}. We also report results on more generic visual reasoning tasks including LogicVista \cite{xiao2024logicvista} and MMMU-Pro\cite{yue2025mmmu}. To ensure a fair comparison, we adopted the exact evaluation setup of VPPO which asks the model to provide answers in \texttt{\textbackslash boxed\{\}} and does not employ and LLM-as-the-Judge for reproducibility (Table \ref{tab:mm_results}).

GCPO outperforms baselines include GRPO, VPPO, and DAPO on both math reasoning tasks and generic visual reasoning tasks. This performance is consistent across different base models. Comparing with DAPO, GCPO exhibits considerable improvements across all tasks. Comparing with VPPO, the improvements is more pronounced in generic visual reasoning tasks such as LogicVista and MMMU-Pro, indicating the VPPO approach which solely filters token gradients based on visual dependencies is suboptimal for tasks also requires strong logical reasoning and world knowledge. By contrast, GCPO assigns continuous weights based on contrastive guidance can better locate key tokens and facilitate the training process. These experiments demonstrated that while GCPO is motivated by CFG in text-to-image inference, it can generalized to domains where CFG is not explicitly used during sampling. 

\subsection{Ablation Studies}

To validate the design choices of GCPO, we conducted extensive ablation studies.

\textbf{Choice of Divergence Metric.} We explored various choices of divergence metrics beyond KL divergence, such as information gain and absolute differences defined below:
\begin{align}
    \text{IG}=\log\frac{\pi_{\theta} (y | x)}{\pi_{\theta} (y | x^-)}, \text{Abs}=|\pi_{\theta} (y | x)-\log \pi_{\theta} (y | x^-)|
\end{align}
Information gain (IG) indicates how much the per-token confidence is increased by the positive prompt. In the context of multimodal understanding with the negative prompt ``generate the wrong answer", it is proportional to the odds of an implicit classifier discussed in equation \ref{eq:odds}. However, unlike KL divergence, IG assigns high weights only to tokens that are likely associated with a correct answer based on the model's belief, while KL also emphasis tokens that are likely associated with an in correct answer. The absolute difference behaves similar to KL divergence in this respect. We report results in Table \ref{tab:divergence_metric}. We find that KL divergence works the best empirically.

\begin{table}[t]
\centering
\caption{\textbf{Comparison of scoring and normalization methods on GenEval and MM12k.}}
\label{tab:ablatio_v1}
\begin{subtable}{0.45\linewidth}
\centering
\caption{Divergence Metric}
\label{tab:divergence_metric}
\begin{tabular}{lcc}
\toprule
 & GenEval & MM12k \\
\midrule
Information Gain & 0.88 & 82.9 \\
Abs difference   & 0.84 & 75.5 \\
KL               & 0.89 & 83.1 \\
\bottomrule
\end{tabular}
\end{subtable}
\hfill
\begin{subtable}{0.45\linewidth}
\centering
\caption{Normalization methods}
\label{tab:normalization}
\begin{tabular}{lcc}
\toprule
 & GenEval & MM12k \\
\midrule
Softmax   & 0.86 & 81.0 \\
Min-max   & 0.85 & 62.5 \\
Histogram & 0.89 & 83.1 \\
\bottomrule
\end{tabular}
\end{subtable}
\end{table}

\begin{table}[t]
\centering
\caption{\textbf{Effect of different negative prompts on MM12k. } $\dagger$ cited from VPPO  \cite{huang2025spotlight}.}
\label{tab:neg_prompt}
\begin{tabular}{lc}
\toprule
Negative Prompt & MM12k \\
\midrule
No Prompt & 76.9 \\
No Visual Input & 81.4 \\
Blank Image $\dagger$& 82.1 \\
Corrupted Visual Input (VPPO)  $\dagger$ & 82.8 \\
Null Prompt ``Answer the question'' & 82.3 \\
Negative suffix ``Generate the wrong answer'' & 83.1 \\
\bottomrule
\end{tabular}
\end{table}

\textbf{Normalization Method} While Figure \ref{fig:normalization} demonstrates the benefits of our proposed normalization technique, we further validate its effectiveness via experiments. We report these results in Table \ref{tab:normalization}, which indicates that our proposed histogram-equalization-style normalization works best or both text to image generation tasks and multimodal understanding tasks. 

\textbf{Choice of Negative Prompts} An important design choice in extending GCPO to multimodal reasoning tasks is the choice of negative prompts, which are not naturally present during rollout. We explored multiple options, including empty string, removing input image tokens, a generic prompt "answer the question", and the negative prompt constructed by adding the suffix "generate the wrong answer" to the original prompt. We also refer to prior experiments from VPPO \cite{huang2025spotlight}, which explored using a blank image and randomly masking the input image. We show these results in Table \ref{tab:neg_prompt}. Our proposed negative suffix design works the best.

\section{Conclusion}

In conclusion, we propose Guidance Contrastive Policy Optimization (GCPO). Unlike GRPO and DAPO which broadcast sample-level advantages uniformly to each token, we make use of the classifier-free guidance signals in text-to-image inference to provide per-token advantages and emphasize learning signals on important regions. We further extend GCPO to text generation, where CFG is not used during rollouts, by designing a negative prompt that reveals the model's implicit belief of token correctness. Extensive experiments demonstrate that GCPO is a generalizable and effective method to assign per-token credits when only sample-level reward is available, paving the way for future works to further advance discrete policy optimization.

\clearpage

\bibliographystyle{plain}
\bibliography{ref}

@article{zheng2025group,
  title={Group sequence policy optimization},
  author={Zheng, Chujie and Liu, Shixuan and Li, Mingze and Chen, Xiong-Hui and Yu, Bowen and Gao, Chang and Dang, Kai and Liu, Yuqiong and Men, Rui and Yang, An and others},
  journal={arXiv preprint arXiv:2507.18071},
  year={2025}
}

@article{wang2025vl,
  title={Vl-rethinker: Incentivizing self-reflection of vision-language models with reinforcement learning},
  author={Wang, Haozhe and Qu, Chao and Huang, Zuming and Chu, Wei and Lin, Fangzhen and Chen, Wenhu},
  journal={arXiv preprint arXiv:2504.08837},
  year={2025}
}

@article{yu2025perception,
  title={Perception-r1: Pioneering perception policy with reinforcement learning},
  author={Yu, En and Lin, Kangheng and Zhao, Liang and Yin, Jisheng and Wei, Yana and Peng, Yuang and Wei, Haoran and Sun, Jianjian and Han, Chunrui and Ge, Zheng and others},
  journal={arXiv preprint arXiv:2504.07954},
  year={2025}
}

@article{shen2025vlm,
  title={Vlm-r1: A stable and generalizable r1-style large vision-language model},
  author={Shen, Haozhan and Liu, Peng and Li, Jingcheng and Fang, Chunxin and Ma, Yibo and Liao, Jiajia and Shen, Qiaoli and Zhang, Zilun and Zhao, Kangjia and Zhang, Qianqian and others},
  journal={arXiv preprint arXiv:2504.07615},
  year={2025}
}

@article{meng2025mm,
  title={Mm-eureka: Exploring the frontiers of multimodal reasoning with rule-based reinforcement learning},
  author={Meng, Fanqing and Du, Lingxiao and Liu, Zongkai and Zhou, Zhixiang and Lu, Quanfeng and Fu, Daocheng and Han, Tiancheng and Shi, Botian and Wang, Wenhai and He, Junjun and others},
  journal={arXiv preprint arXiv:2503.07365},
  year={2025}
}

@article{liu2025flow,
  title={Flow-grpo: Training flow matching models via online rl},
  author={Liu, Jie and Liu, Gongye and Liang, Jiajun and Li, Yangguang and Liu, Jiaheng and Wang, Xintao and Wan, Pengfei and Zhang, Di and Ouyang, Wanli},
  journal={arXiv preprint arXiv:2505.05470},
  year={2025}
}

@article{schulman2017proximal,
  title={Proximal policy optimization algorithms},
  author={Schulman, John and Wolski, Filip and Dhariwal, Prafulla and Radford, Alec and Klimov, Oleg},
  journal={arXiv preprint arXiv:1707.06347},
  year={2017}
}

@article{wu2025qwen,
  title={Qwen-image technical report},
  author={Wu, Chenfei and Li, Jiahao and Zhou, Jingren and Lin, Junyang and Gao, Kaiyuan and Yan, Kun and Yin, Sheng-ming and Bai, Shuai and Xu, Xiao and Chen, Yilei and others},
  journal={arXiv preprint arXiv:2508.02324},
  year={2025}
}

@article{deepseek2025r1,
  title={DeepSeek-R1: Incentivizing Reasoning Capability in LLMs via Reinforcement Learning},
  author={DeepSeek-AI},
  journal={arXiv preprint arXiv:2501.12948},
  year={2025}
}

@article{zheng2025diffusionnft,
  title={DiffusionNFT: Online Diffusion Reinforcement with Forward Process},
  author={Zheng, Kaiwen and Chen, Huayu and Ye, Haotian and Wang, Haoxiang and Zhang, Qinsheng and Jiang, Kai and Su, Hang and Ermon, Stefano and Zhu, Jun and Liu, Ming-Yu},
  journal={arXiv preprint arXiv:2509.16117},
  year={2025}
}

@article{rafailov2023direct,
  title={Direct preference optimization: Your language model is secretly a reward model},
  author={Rafailov, Rafael and Sharma, Archit and Mitchell, Eric and Manning, Christopher D and Ermon, Stefano and Finn, Chelsea},
  journal={Advances in neural information processing systems},
  volume={36},
  pages={53728--53741},
  year={2023}
}

@article{wei2022chain,
  title={Chain-of-thought prompting elicits reasoning in large language models},
  author={Wei, Jason and Wang, Xuezhi and Schuurmans, Dale and Bosma, Maarten and Xia, Fei and Chi, Ed and Le, Quoc V and Zhou, Denny and others},
  journal={Advances in neural information processing systems},
  volume={35},
  pages={24824--24837},
  year={2022}
}

@article{schulman2016gae,
  title={High-Dimensional Continuous Control Using Generalized Advantage Estimation},
  author={Schulman, John and Moritz, Philipp and Levine, Sergey and Jordan, Michael I. and Abbeel, Pieter},
  journal={arXiv preprint arXiv:1506.02438},
  year={2016}
}

@article{christiano2017deep,
  title={Deep Reinforcement Learning from Human Preferences},
  author={Christiano, Paul F. and Leike, Jan and Brown, Tom B. and Martic, Miljan and Legg, Shane and Amodei, Dario},
  journal={NeurIPS},
  year={2017}
}

@article{o2023contrastive,
  title={Contrastive decoding improves reasoning in large language models},
  author={O'Brien, Sean and Lewis, Mike},
  journal={arXiv preprint arXiv:2309.09117},
  year={2023}
}

@article{oneill2023steer,
  title={Steering Language Generation: Harnessing Contrastive Expert Guidance and Negative Prompting},
  author={O'Neill, Charles and others},
  journal={arXiv preprint arXiv:2308.07645},
  year={2023}
}

@article{nichol2021glide,
  title={GLIDE: Towards Photorealistic Image Generation and Editing with Text-Guided Diffusion Models},
  author={Nichol, Alex and Dhariwal, Prafulla},
  journal={arXiv preprint arXiv:2112.10741},
  year={2021}
}

@article{saharia2022imagen,
  title={Photorealistic Text-to-Image Diffusion Models with Deep Language Understanding},
  author={Saharia, Chitwan and others},
  journal={arXiv preprint arXiv:2205.11487},
  year={2022}
}

@article{rombach2022highresolution,
  title={High-Resolution Image Synthesis with Latent Diffusion Models},
  author={Rombach, Robin and Blattmann, Andreas and Lorenz, Dominik and Esser, Patrick and Ommer, Björn},
  journal={arXiv preprint arXiv:2112.10752},
  year={2022}
}

@article{bai2025qwen25-vl,
  title={Qwen2. 5-vl technical report},
  author={Bai, Shuai and Chen, Keqin and Liu, Xuejing and Wang, Jialin and Ge, Wenbin and Song, Sibo and Dang, Kai and Wang, Peng and Wang, Shijie and Tang, Jun and others},
  journal={arXiv preprint arXiv:2502.13923},
  year={2025}
}

@inproceedings{esser2024scaling-sd3,
  title={Scaling rectified flow transformers for high-resolution image synthesis},
  author={Esser, Patrick and Kulal, Sumith and Blattmann, Andreas and Entezari, Rahim and M{\"u}ller, Jonas and Saini, Harry and Levi, Yam and Lorenz, Dominik and Sauer, Axel and Boesel, Frederic and others},
  booktitle={Forty-first international conference on machine learning},
  year={2024}
}

@article{zhang2024mathverse,
  title={MathVerse: Does Your Multi-modal LLM Truly See the Diagrams in Visual Math Problems?},
  author={Zhang, Renrui and Jiang, Dongzhi and Zhang, Yichi and Lin, Haokun and Guo, Ziyu and Qiu, Pengshuo and Zhou, Aojun and Lu, Pan and Chang, Kai-Wei and Gao, Peng and others},
  journal={arXiv preprint arXiv:2403.14624},
  year={2024}
}

@inproceedings{
    wang2024measuring,
    title={Measuring Multimodal Mathematical Reasoning with MATH-Vision Dataset},
    author={Ke Wang and Junting Pan and Weikang Shi and Zimu Lu and Houxing Ren and Aojun Zhou and Mingjie Zhan and Hongsheng Li},
    booktitle={The Thirty-eight Conference on Neural Information Processing Systems Datasets and Benchmarks Track},
    year={2024},
    url={https://openreview.net/forum?id=QWTCcxMpPA}
}

@article{yang2025mmada,
  title   = {Multimodal Large Diffusion Language Models},
  author  = {Yang, Ling and Tian, Ye and Li, Bowen and Zhang, Xinchen and Shen, Ke and Tong, Yunhai and Wang, Mengdi},
  journal = {arXiv preprint arXiv:2505.15809},
  year    = {2025}
}

@article{deng2025emerging,
  title={Emerging properties in unified multimodal pretraining},
  author={Deng, Chaorui and Zhu, Deyao and Li, Kunchang and Gou, Chenhui and Li, Feng and Wang, Zeyu and Zhong, Shu and Yu, Weihao and Nie, Xiaonan and Song, Ziang and others},
  journal={arXiv preprint arXiv:2505.14683},
  year={2025}
}

@article{chen2025janus,
  title={Janus-pro: Unified multimodal understanding and generation with data and model scaling},
  author={Chen, Xiaokang and Wu, Zhiyu and Liu, Xingchao and Pan, Zizheng and Liu, Wen and Xie, Zhenda and Yu, Xingkai and Ruan, Chong},
  journal={arXiv preprint arXiv:2501.17811},
  year={2025}
}

@misc{flux2024,
    author={Black Forest Labs},
    title={FLUX},
    year={2024},
    howpublished={\url{https://github.com/black-forest-labs/flux}},
}

@article{ye2025data,
  title={Data-regularized Reinforcement Learning for Diffusion Models at Scale},
  author={Ye, Haotian and Zheng, Kaiwen and Xu, Jiashu and Li, Puheng and Chen, Huayu and Han, Jiaqi and Liu, Sheng and Zhang, Qinsheng and Mao, Hanzi and Hao, Zekun and others},
  journal={arXiv preprint arXiv:2512.04332},
  year={2025}
}

@article{Wallace2023DiffusionMA,
  title={Diffusion Model Alignment Using Direct Preference Optimization},
  author={Bram Wallace and Meihua Dang and Rafael Rafailov and Linqi Zhou and Aaron Lou and Senthil Purushwalkam and Stefano Ermon and Caiming Xiong and Shafiq R. Joty and Nikhil Naik},
  journal={2024 IEEE/CVF Conference on Computer Vision and Pattern Recognition (CVPR)},
  year={2023},
  pages={8228-8238},
  url={https://api.semanticscholar.org/CorpusID:265352136}
}

@inproceedings{ban2024understanding,
  title={Understanding the impact of negative prompts: When and how do they take effect?},
  author={Ban, Yuanhao and Wang, Ruochen and Zhou, Tianyi and Cheng, Minhao and Gong, Boqing and Hsieh, Cho-Jui},
  booktitle={european conference on computer vision},
  pages={190--206},
  year={2024},
  organization={Springer}
}

@article{stiennon2020learning,
  title={Learning to Summarize with Human Feedback},
  author={Stiennon, Nisan and others},
  journal={NeurIPS},
  year={2020}
}

@article{ghosh2023geneval,
  title={Geneval: An object-focused framework for evaluating text-to-image alignment},
  author={Ghosh, Dhruba and Hajishirzi, Hannaneh and Schmidt, Ludwig},
  journal={Advances in Neural Information Processing Systems},
  volume={36},
  pages={52132--52152},
  year={2023}
}

@misc{openai_dalle3,
  author       = {OpenAI},
  title        = {DALL·E 3},
  year         = {2023},
  howpublished = {\url{https://openai.com/index/dall-e-3/}},
}

@article{li2025lavidao,
  title={Lavida-O: Elastic Masked Diffusion Models for Unified Multimodal Understanding and Generation},
  author={Li, Shufan and Gu, Jiuxiang and Liu, Kangning and Lin, Zhe and Wei, Zijun and Grover, Aditya and Kuen, Jason},
  journal={arXiv preprint arXiv:2509.19244},
  year={2025}
}

@inproceedings{yue2025mmmu,
  title={Mmmu-pro: A more robust multi-discipline multimodal understanding benchmark},
  author={Yue, Xiang and Zheng, Tianyu and Ni, Yuansheng and Wang, Yubo and Zhang, Kai and Tong, Shengbang and Sun, Yuxuan and Yu, Botao and Zhang, Ge and Sun, Huan and others},
  booktitle={Proceedings of the 63rd Annual Meeting of the Association for Computational Linguistics (Volume 1: Long Papers)},
  pages={15134--15186},
  year={2025}
}

@article{shao2024deepseekmath,
  title={Deepseekmath: Pushing the limits of mathematical reasoning in open language models},
  author={Shao, Zhihong and Wang, Peiyi and Zhu, Qihao and Xu, Runxin and Song, Junxiao and Bi, Xiao and Zhang, Haowei and Zhang, Mingchuan and Li, YK and Wu, Yang and others},
  journal={arXiv preprint arXiv:2402.03300},
  year={2024}
}

@article{Qwen3-VL,
      title={Qwen3-VL Technical Report}, 
      author={Shuai Bai and Yuxuan Cai and Ruizhe Chen and Keqin Chen and Xionghui Chen and Zesen Cheng and Lianghao Deng and Wei Ding and Chang Gao and Chunjiang Ge and Wenbin Ge and Zhifang Guo and Qidong Huang and Jie Huang and Fei Huang and Binyuan Hui and Shutong Jiang and Zhaohai Li and Mingsheng Li and Mei Li and Kaixin Li and Zicheng Lin and Junyang Lin and Xuejing Liu and Jiawei Liu and Chenglong Liu and Yang Liu and Dayiheng Liu and Shixuan Liu and Dunjie Lu and Ruilin Luo and Chenxu Lv and Rui Men and Lingchen Meng and Xuancheng Ren and Xingzhang Ren and Sibo Song and Yuchong Sun and Jun Tang and Jianhong Tu and Jianqiang Wan and Peng Wang and Pengfei Wang and Qiuyue Wang and Yuxuan Wang and Tianbao Xie and Yiheng Xu and Haiyang Xu and Jin Xu and Zhibo Yang and Mingkun Yang and Jianxin Yang and An Yang and Bowen Yu and Fei Zhang and Hang Zhang and Xi Zhang and Bo Zheng and Humen Zhong and Jingren Zhou and Fan Zhou and Jing Zhou and Yuanzhi Zhu and Ke Zhu},
	  journal={arXiv preprint arXiv:2511.21631},
      year={2025}
}

@article{ouyang2022training,
  title={Training language models to follow instructions with human feedback},
  author={Ouyang, Long and Wu, Jeffrey and Jiang, Xu and Almeida, Diogo and Wainwright, Carroll and Mishkin, Pamela and Zhang, Chong and Agarwal, Sandhini and Slama, Katarina and Ray, Alex and others},
  journal={Advances in neural information processing systems},
  volume={35},
  pages={27730--27744},
  year={2022}
}

@article{pan2025janus,
  title={Janus-pro-r1: Advancing collaborative visual comprehension and generation via reinforcement learning},
  author={Pan, Kaihang and Wu, Yang and Bu, Wendong and Shen, Kai and Li, Juncheng and Wang, Yingting and Li, Yunfei and Tang, Siliang and Xiao, Jun and Wu, Fei and others},
  journal={arXiv preprint arXiv:2506.01480},
  year={2025}
}

@article{yu2025dapo,
  title={Dapo: An open-source llm reinforcement learning system at scale},
  author={Yu, Qiying and Zhang, Zheng and Zhu, Ruofei and Yuan, Yufeng and Zuo, Xiaochen and Yue, Yu and Dai, Weinan and Fan, Tiantian and Liu, Gaohong and Liu, Lingjun and others},
  journal={arXiv preprint arXiv:2503.14476},
  year={2025}
}

@article{liu2025understanding,
  title={Understanding r1-zero-like training: A critical perspective},
  author={Liu, Zichen and Chen, Changyu and Li, Wenjun and Qi, Penghui and Pang, Tianyu and Du, Chao and Lee, Wee Sun and Lin, Min},
  journal={arXiv preprint arXiv:2503.20783},
  year={2025}
}

@article{xie2025unlocking,
  title={Unlocking exploration in rlvr: Uncertainty-aware advantage shaping for deeper reasoning},
  author={Xie, Can and Pan, Ruotong and Wu, Xiangyu and Zhang, Yunfei and Fu, Jiayi and Gao, Tingting and Zhou, Guorui},
  journal={arXiv preprint arXiv:2510.10649},
  year={2025}
}

@article{wang2025beyond,
  title={Beyond the 80/20 rule: High-entropy minority tokens drive effective reinforcement learning for llm reasoning},
  author={Wang, Shenzhi and Yu, Le and Gao, Chang and Zheng, Chujie and Liu, Shixuan and Lu, Rui and Dang, Kai and Chen, Xionghui and Yang, Jianxin and Zhang, Zhenru and others},
  journal={arXiv preprint arXiv:2506.01939},
  year={2025}
}

@article{sun2025ktae,
  title={KTAE: A Model-Free Algorithm to Key-Tokens Advantage Estimation in Mathematical Reasoning},
  author={Sun, Wei and Yang, Wen and Jian, Pu and Du, Qianlong and Cui, Fuwei and Ren, Shuo and Zhang, Jiajun},
  journal={arXiv preprint arXiv:2505.16826},
  year={2025}
}

@article{huang2025spotlight,
  title={Spotlight on token perception for multimodal reinforcement learning},
  author={Huang, Siyuan and Qu, Xiaoye and Li, Yafu and Luo, Yun and He, Zefeng and Liu, Daizong and Cheng, Yu},
  journal={arXiv preprint arXiv:2510.09285},
  year={2025}
}

@article{ho2022classifier,
  title={Classifier-free diffusion guidance},
  author={Ho, Jonathan and Salimans, Tim},
  journal={arXiv preprint arXiv:2207.12598},
  year={2022}
}

@article{sanchez2023stay,
  title={Stay on topic with classifier-free guidance},
  author={Sanchez, Guillaume and Fan, Honglu and Spangher, Alexander and Levi, Elad and Ammanamanchi, Pawan Sasanka and Biderman, Stella},
  journal={arXiv preprint arXiv:2306.17806},
  year={2023}
}

@article{xiao2024logicvista,
  title={Logicvista: Multimodal llm logical reasoning benchmark in visual contexts},
  author={Xiao, Yijia and Sun, Edward and Liu, Tianyu and Wang, Wei},
  journal={arXiv preprint arXiv:2407.04973},
  year={2024}
}

@article{wang2024emu3,
  title={Emu3: Next-token prediction is all you need},
  author={Wang, Xinlong and Zhang, Xiaosong and Luo, Zhengxiong and Sun, Quan and Cui, Yufeng and Wang, Jinsheng and Zhang, Fan and Wang, Yueze and Li, Zhen and Yu, Qiying and others},
  journal={arXiv preprint arXiv:2409.18869},
  year={2024}
}

@article{chen2023pixart,
  title={Pixart-$\alpha$: Fast training of diffusion transformer for photorealistic text-to-image synthesis},
  author={Chen, Junsong and Yu, Jincheng and Ge, Chongjian and Yao, Lewei and Xie, Enze and Wu, Yue and Wang, Zhongdao and Kwok, James and Luo, Ping and Lu, Huchuan and others},
  journal={arXiv preprint arXiv:2310.00426},
  year={2023}
}

@article{xie2025show,
  title={Show-o2: Improved native unified multimodal models},
  author={Xie, Jinheng and Yang, Zhenheng and Shou, Mike Zheng},
  journal={arXiv preprint arXiv:2506.15564},
  year={2025}
}

\newpage


\appendix

\section{Additional Implementation Details}

\textbf{Text-to-Image.} We employ Janus-Pro-7B as the base model and train GCPO for 1,600 steps. We employ AdamW optimizer with a cosine decay learning rate. The specific hyperparameter are listed in Table \ref{tab:hp1}

\begin{table}[h]
\centering
\caption{\textbf{Training configurations for Text-to-Image tasks.}}
\label{tab:hp1}
\begin{tabular}{lcHH}
\toprule

Learning Rate & $3 \times 10^{-6}$  & $1 \times 10^{-5}$ & $2 \times 10^{-5}$ \\
Steps & 1600 & 100k & 100k \\
$\beta_1$ & 0.99 &0.99  & 0.99 \\
$\beta_2$ & 0.999 &0.999  & 0.999 \\
optimizer & AdamW & AdamW & AdamW \\
Learning Rate Schedule & Cosine  & Cosine & Cosine \\
\midrule
Model Size & 8B & 8B & 10.4B \\
Global Batch Size & 128 & 256 \\
Group Size & 16 & 256 \\
KL regularization weight $\beta$ & 0.03 & \\
CFG scale & 5.0 \\
\bottomrule
\end{tabular}%

\end{table}

\textbf{Multimodal Understanding.} We employ Qwen-2.5-VL-7B-Instruct \cite{bai2025qwen25-vl} and Qwen-3-VL-8B-Instruct \cite{Qwen3-VL} as our base model. We followed the setup of VPPO and train the model for two epochs, which amounts to 202 steps on ViRL39K dataset. The specific hyperparameter are listed in Table \ref{tab:hp2}

\begin{table}[h]
\centering
\caption{\textbf{Training configurations for Text-to-Image tasks.}}
\label{tab:hp2}
\begin{tabular}{lcHH}
\toprule

Learning Rate & $1 \times 10^{-6}$  & $1 \times 10^{-5}$ & $2 \times 10^{-5}$ \\
Steps & 202 & 100k & 100k \\
$\beta_1$ & 0.99 &0.99  & 0.99 \\
$\beta_2$ & 0.999 &0.999  & 0.999 \\
optimizer & AdamW & AdamW & AdamW \\
Learning Rate Schedule & Constant  & Cosine & Cosine \\
\midrule
Model Size & 8B & 8B & 10.4B \\
Global Batch Size & 130 & 256 \\
Group Size & 8 & 256 \\
KL regularization weight $\beta$ & 0 & \\
\bottomrule
\end{tabular}%

\end{table}

During evaluation, we followed VPPO and sample 8 responses per question and report the average accuracy.

\section{Training Dynamics}

\begin{figure}[t]
    \centering
    \includegraphics[width=0.6\linewidth]{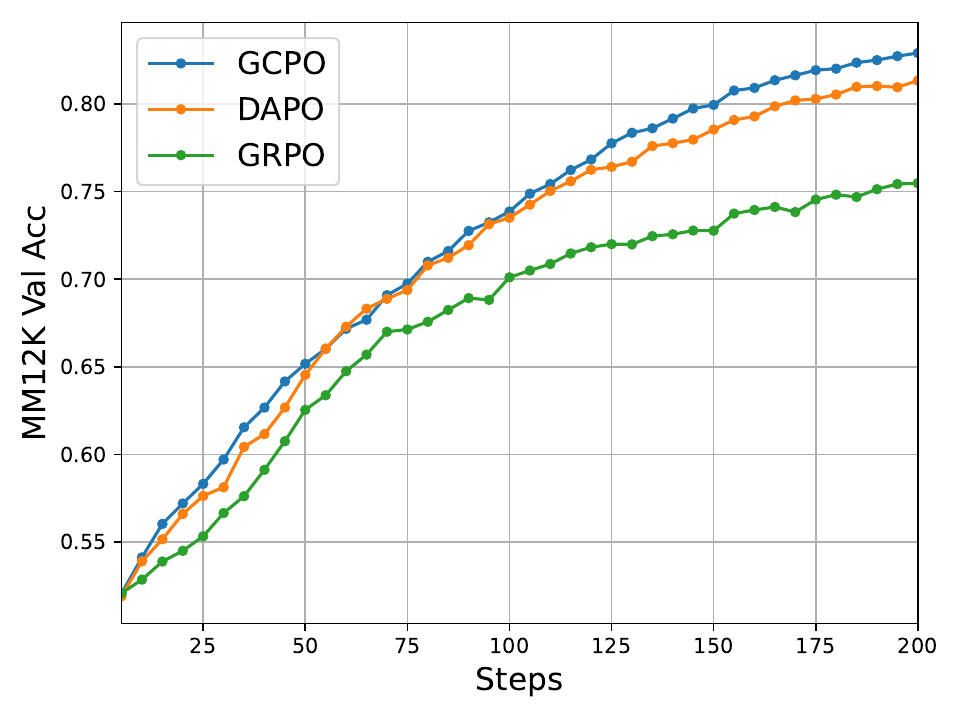}
    \caption{\textbf{Training Dynamics of GCPO.} We visualize the progression of MM12K validation accuracy at different training steps for GCPO, DAPO, and GRPO experiments.}
    \label{fig:reward_curve}
\end{figure}

We also visualize the validation reward curve of GCPO, DAPO, and GRPO in Figure \ref{fig:reward_curve}. These results show that GCPO consistently outperforms DAPO at most training stages, with the performance gap growing bigger as the training progresses.

\textbf{Compute Usage} For all models we train with 8 B200 GPUs. The training time for text-to-image experiments is 30 hours while the training time for multimodal understanding and reasoning tasks takes 40 hours. Notably, we report Avg\@8 for multi model reasoning tasks, which takes 6 hours per evaluation run using 8 GPUs because model generate long responses.

\section{Visual Examination of Contrastive Guidance}

\begin{figure}
    \centering
    \includegraphics[width=0.8\linewidth]{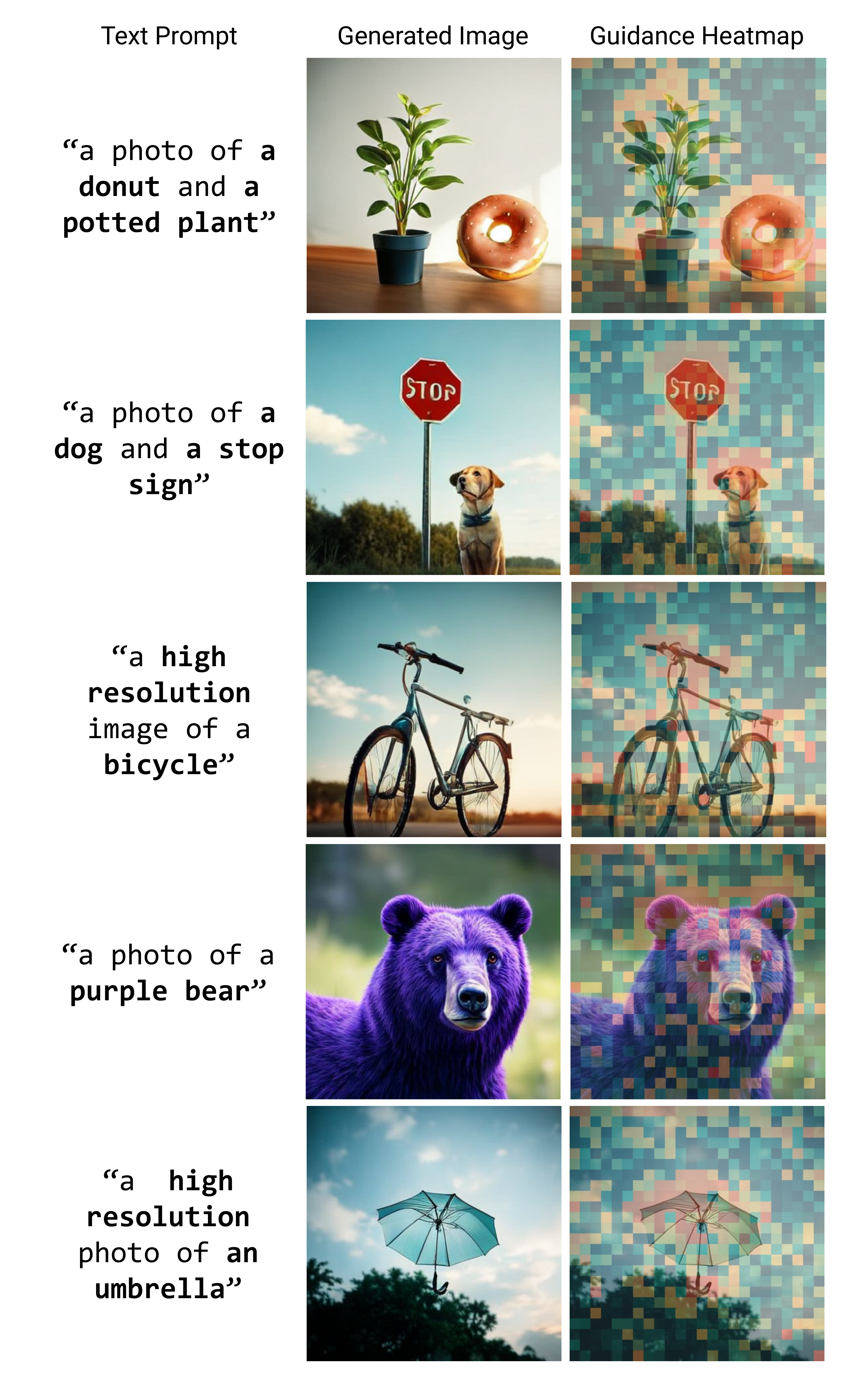}
    \caption{\textbf{Qualitative Examples of Heatmap Guidance}}
    \label{fig:heatmap_app}
\end{figure}

We provide additional visualizations of contrastive guidance for text-to-image tasks in figure \ref{fig:heatmap_app} and multimodal understanding tasks in figure \ref{fig:mm_heatmap}. These results demonstrate that GCPO effectively focus the learning signal to critical regions and tokens.

\section{Qualitative Results for multimodal reasoning. }

\begin{figure}
    \centering
    \includegraphics[width=1.0\linewidth]{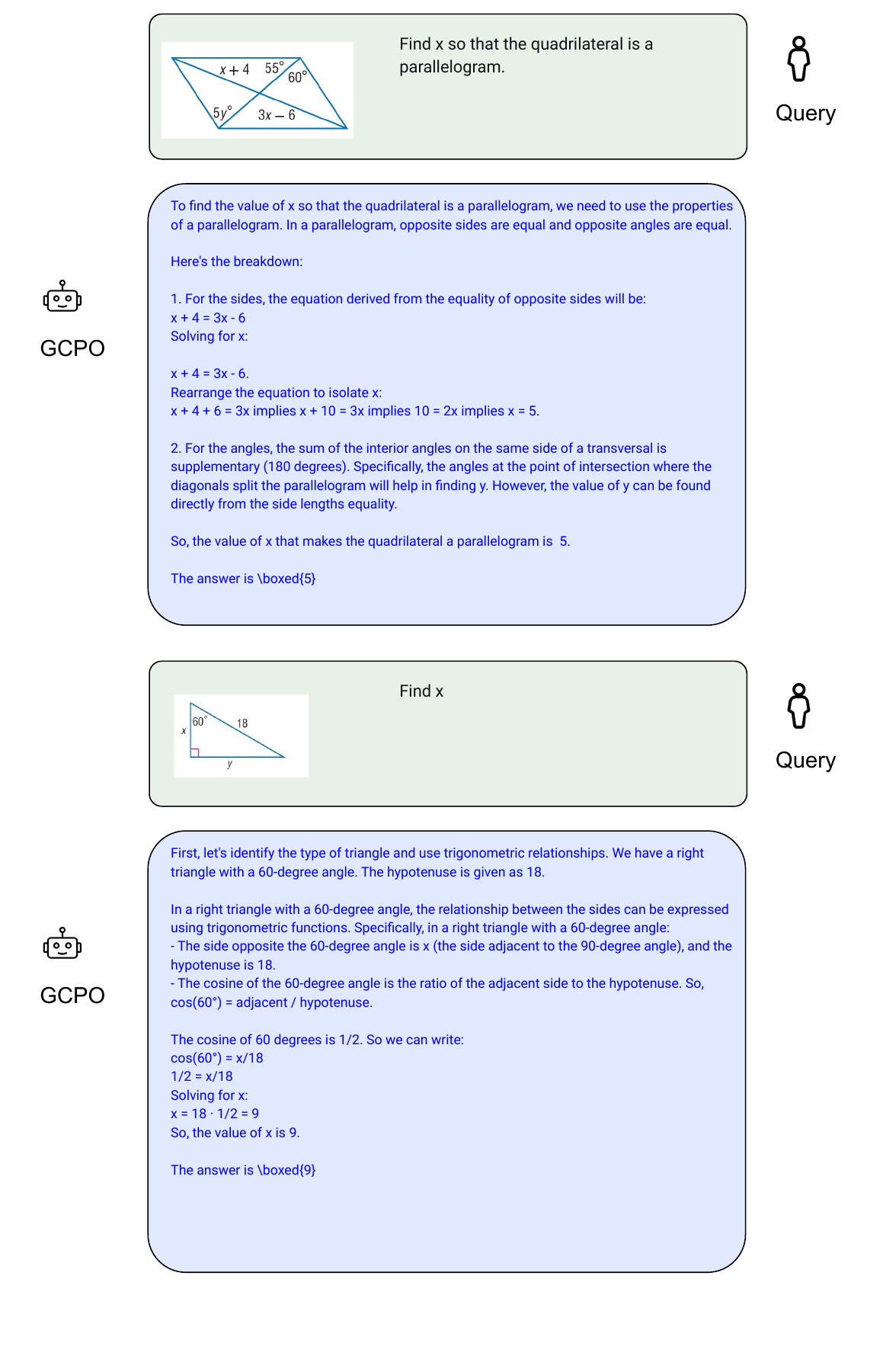}
    \caption{\textbf{Qualitative Examples of Multimodal Reasoning (1/3).}}
    \label{fig:demo_mm_1}
\end{figure}

\begin{figure}
    \centering
    \includegraphics[width=1.0\linewidth]{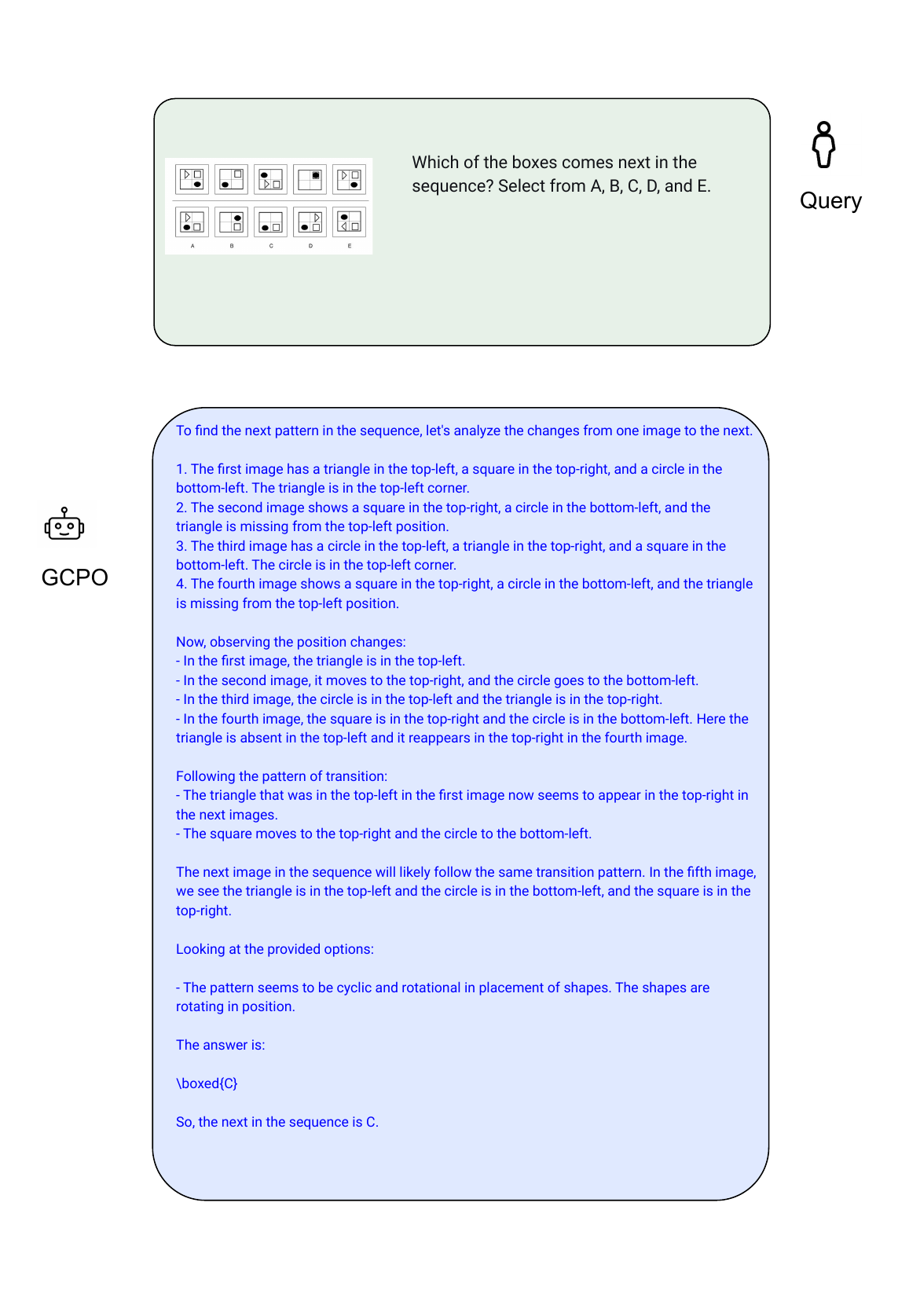}
    \caption{\textbf{Qualitative Examples of Multimodal Reasoning (2/3).}}
    \label{fig:demo_mm_2}
\end{figure}

\begin{figure}
    \centering
    \includegraphics[width=1.0\linewidth]{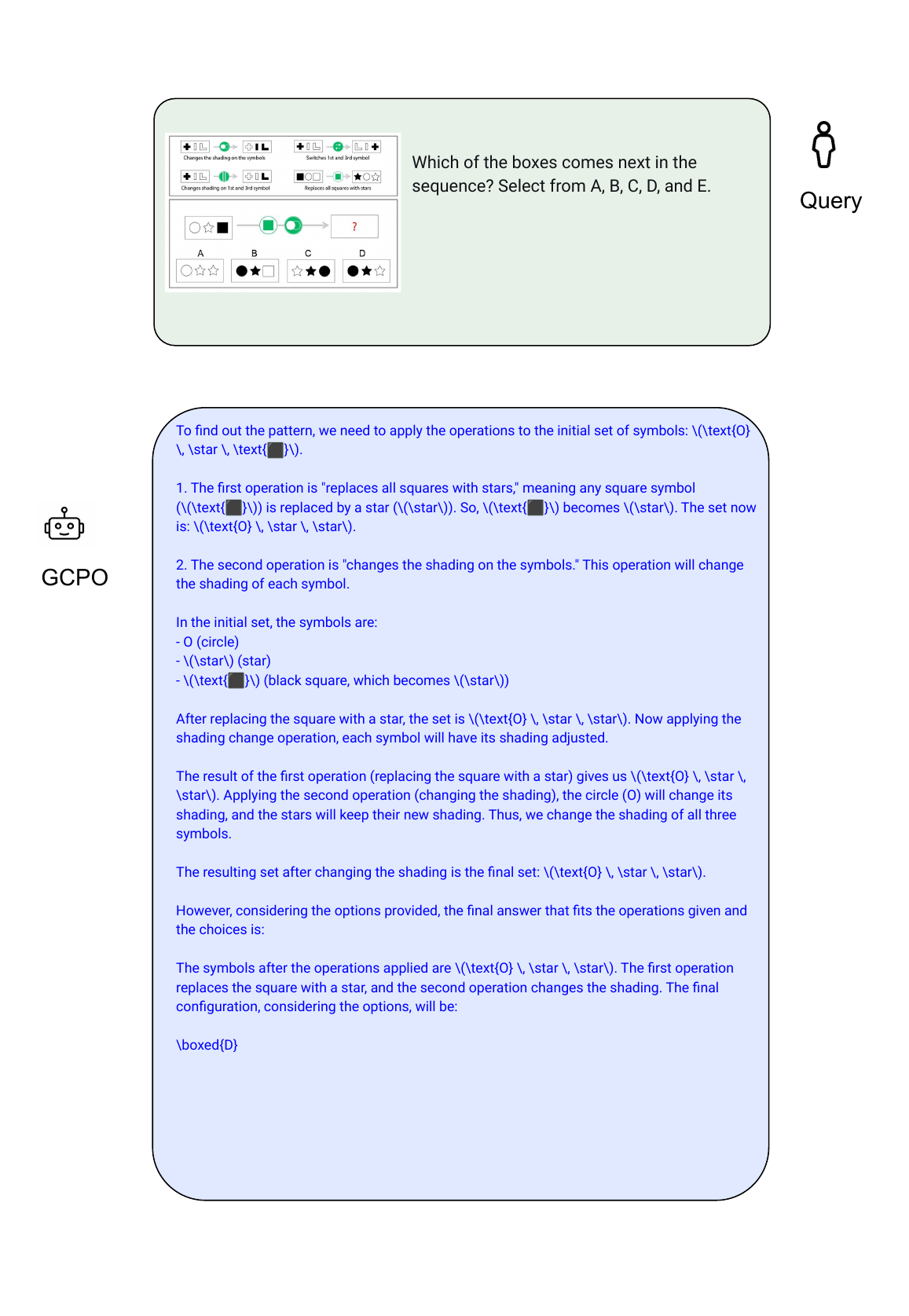}
    \caption{\textbf{Qualitative Examples of Multimodal Reasoning (3/3).}}
    \label{fig:demo_mm_3}
\end{figure}

We provide output samples of GCPO-finetuned Qwen2.5-VL-7B-Instruct model in figure \ref{fig:demo_mm_1}, figure \ref{fig:demo_mm_2}, and figure \ref{fig:demo_mm_3}, highlighting the strong multimodal reasoning capabilities of GCPO-enhanced VLMs.

\begin{figure}
    \centering
    \includegraphics[width=1.0\linewidth]{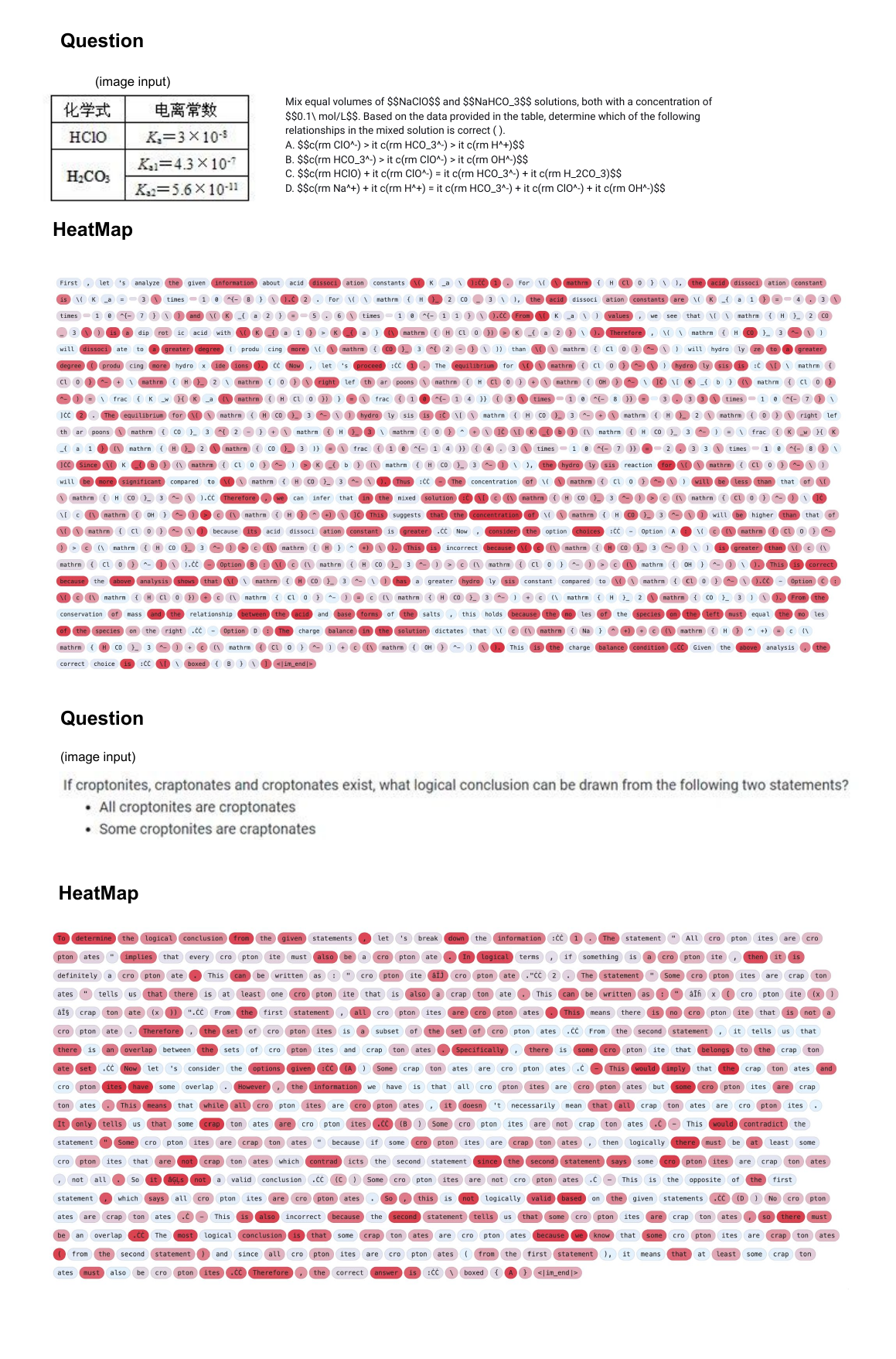}
    \caption{\textbf{Visualizations of per token weighting under GCPO framework.} Darker colors indicates higher weighting. }
    \label{fig:mm_heatmap}
\end{figure}

\begin{figure}
    \centering
    \includegraphics[width=1.0\linewidth]{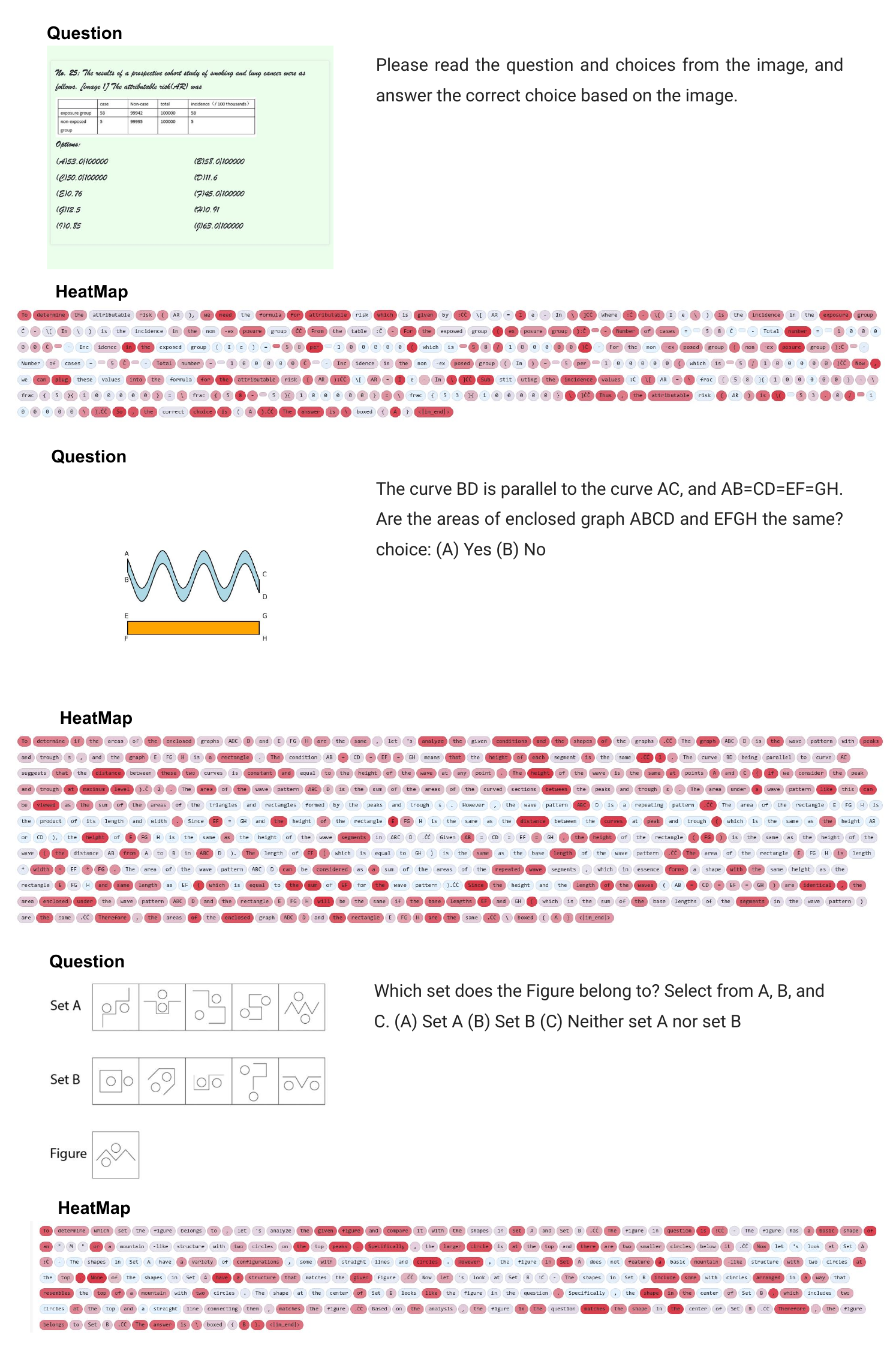}
    \caption{\textbf{More visualization of per token weighting under GCPO framework.} Darker colors indicates higher weighting. }
    \label{fig:mm_heatmap}
\end{figure}

\section{Limitation}

\label{sec:limitation}

Despite the strong results of \ours, it has two key limitations. First, it only works for reinforcement learning tasks that are strictly prompt dependent, which is the precondition for GCPO to derive contrastive guidance with positive prompt $x$ and negative prompt $x^-$. For example, it cannot apply to RL setting whose reward is simply the response length, as the token weight scheme of GCPO focuses on key tokens that are critical to the correctness of the response, which is irrelevant in this setting. 

Second, for multimodal understanding tasks it assumes the model have some initial capabilities of differentiating a correct and incorrect answer. We find smaller models at 1B scale may not respond to the instruction ``generate a wrong answer" well and would still generate a correct answer even when prompted not to do so, this makes the guidance an unreliable signal since the model cannot properly comprehend the negative prompts. Employing an external teacher model to provide GCPO-style per token weighting may be feasible, we will explore this setup in future works

\section{Broader Impact}

Our works propose a novel algorithm to finetune generative models. However, when used improperly, it may be used to train a model to perform malicious actions. Even without malicious intent, the finetuned model may still inherent the biases and hallucinations of the base model. We do not recommend non-research used for finetuned models.

\label{sec:impact}
\clearpage
\section{Licenses }

We report the licenses of the used artifacts in Table \ref{tab:license-info}. We followed the intended use of all respective artifacts.

\begin{table}[H]
\centering
\renewcommand{\arraystretch}{1.3}
\setlength{\tabcolsep}{6pt}
\caption{\textbf{Licenses and sources for datasets and models used.} \textsuperscript{†}Code under MIT License; weights under DeepSeek Model License.}
\begin{tabularx}{\textwidth}{lXlX}
\toprule
\textbf{Category} & \textbf{Name} & \textbf{License} & \textbf{Platform} \\
\midrule
Base Model & Janus-Pro-7B & Varies \textsuperscript{†} & Hugging Face \\
Base Model & Janus-Pro-R1 & Varies \textsuperscript{†} & Hugging Face \\
VLM & Qwen2.5-VL-7B-Instruct & Apache 2.0 & Hugging Face \\
VLM & Qwen3-VL-8B-Instruct & Apache 2.0 & Hugging Face \\
\midrule
Train Dataset & FlowGRPO Data & MIT & GitHub \\
Train Dataset & ViRL39K & MIT & Hugging Face \\
\midrule
Eval Benchmark & GenEval & MIT & GitHub \\
Eval Benchmark & MathVerse & MIT & Hugging Face \\
Eval Benchmark & MathVision & MIT & Hugging Face \\
Eval Benchmark & MM12K (MM-Eureka) & Apache 2.0 & GitHub \\
Eval Benchmark & LogicVista & Apache 2.0 & GitHub \\
Eval Benchmark & MMMU-Pro & Apache 2.0 & Hugging Face \\
\bottomrule
\end{tabularx}
\label{tab:license-info}
\end{table}

\newpage

\end{document}